# Color Spike Data Generation via Bio-inspired Neuron-like Encoding with an Artificial Photoreceptor Layer

Ching-Teng Hsieh[1], and Yuan-Kai Wang[12]
[1]Department of Electrical Engineering, Fu Jen Catholic University, New Taipei City 242062, Taiwan
[2]Graduate Institute of Applied Science and Engineering, Fu Jen Catholic University, New Taipei City 242062, Taiwan
Corresponding author: Yuan-Kai, Wang (e-mail: ykwang@fju.edu.tw).

**ABSTRACT** In recent years, neuromorphic computing and spiking neural networks (SNNs) have ad-vanced rapidly through integration with deep learning. However, the performance of SNNs still lags behind that of convolutional neural networks (CNNs), primarily due to the limited information capacity of spike-based data. Although some studies have attempted to improve SNN performance by training them with non-spiking inputs such as static images, this approach deviates from the original intent of neuromorphic computing, which emphasizes spike-based information processing. To address this issue, we propose a Neuron-like Encoding method that generates spike data based on the intrinsic operational principles and functions of biological neurons. This method is further enhanced by the incorporation of an artificial pho-toreceptor layer, enabling spike data to carry both color and luminance information, thereby forming a complete visual spike signal. Experimental results using the Integrate-and-Fire neuron model demonstrate that this biologically inspired approach effectively increases the information content of spike signals and im-proves SNN performance, all while adhering to neuromorphic principles. We believe this concept holds strong potential for future development and may contribute to overcoming current limitations in neuro-morphic computing, facilitating broader applications of SNNs.

## I. INTRODUCTION

Over the past decade, rapid technological advancements—particularly the significant improvement in Graphics Processing Unit (GPU) performance—have accelerated the development of general-purpose computing on graphics processing units (GPGPU). With the rise of GPGPU, the focus of artificial intelligence (AI) research has gradually shifted from traditional machine learning (ML) to deep learning (DL), which employs large datasets to train artificial neural networks (ANNs) with nonlinear characteristics. These models have enabled substantial progress in tasks such as automatic classification and detection. However, the success of deep learning has also brought about considerable power consumption, as floating-point operations in hardware require substantial computational resources.

To address the power inefficiency of ANNs, researchers have turned to neuromorphic computing [1], a paradigm that replaces artificial neurons with spiking neurons and builds spiking neural networks (SNNs). These networks process spike-based or event-driven data in an effort to drastically reduce power consumption. While spike data can be generated via traditional neural coding schemes or captured through event cameras, such data generally lack color information and often miss critical background texture details. This limitation hinders the broad applicability of SNNs and constrains their performance across diverse tasks.

For example, consider a robotic system that must identify and retrieve a cup from a cabinet of a specific color and shape. While replacing conventional cameras with event cameras allows the robot to perceive dynamic changes in the environment and detect cabinet shapes during movement, it becomes problematic when multiple cabinets of identical shape but different colors exist, or when the robot remains stationary and loses visual input. In such cases, the robot cannot determine which cabinet to open or may lose track of the target, entering a failure loop. Despite these limitations, the high spatiotemporal resolution of event cameras remains unmatched by traditional imaging systems, offering the advantage of extremely fine temporal granularity. This allows robotic systems to react faster than those relying solely on RGB imagery.

To address the lack of color awareness in SNNs, recent research efforts [2, 3] have attempted to train SNNs using static images, achieving performance comparable to that of convolutional neural networks (CNNs) and Transformers. However, such approaches diverge from the core philosophy of neuromorphic computing, which emphasizes spike-based and event-driven data as ideal input modalities for achieving ultra-low power consumption.

In response, we begin by re-examining existing neural coding techniques to identify their limitations. Inspired by the structure of the human retina, we propose a novel Neuron-like Encoding method that leverages biologically plausible neuron models to convert static images into spike data, preserving both rate and temporal information. Furthermore, we introduce the concept of an Artificial Visual Photoreceptor Layer, which mimics the function of rod and cone cells in the retina. This layer integrates both color and luminance information into multidimensional spike representations, enabling SNNs to receive rich, near-complete visual signals. The overall system architecture is illustrated in Fig. 1.

We validate our approach by training four different SNN models and evaluating the impact of color information and the proposed encoding method on model performance. Additionally, we explore various color space representations to assess the effect of different embedded features on accuracy. We also analyze the computational efficiency and power consumption of spike-based versus static image-based inputs.

This paper is structured as follows:
- Section 2 provides a comprehensive review of current imaging technologies, neuromorphic computing, SNNs and their input modalities, as well as existing neural coding techniques.
- Section 3 details the proposed Neuron-like Encoding and the Artificial Visual Photoreceptor Layer.
- Section 4 presents experimental results and analysis on the effectiveness of the proposed method.
- Section 5 concludes the study and discusses potential directions for future work.

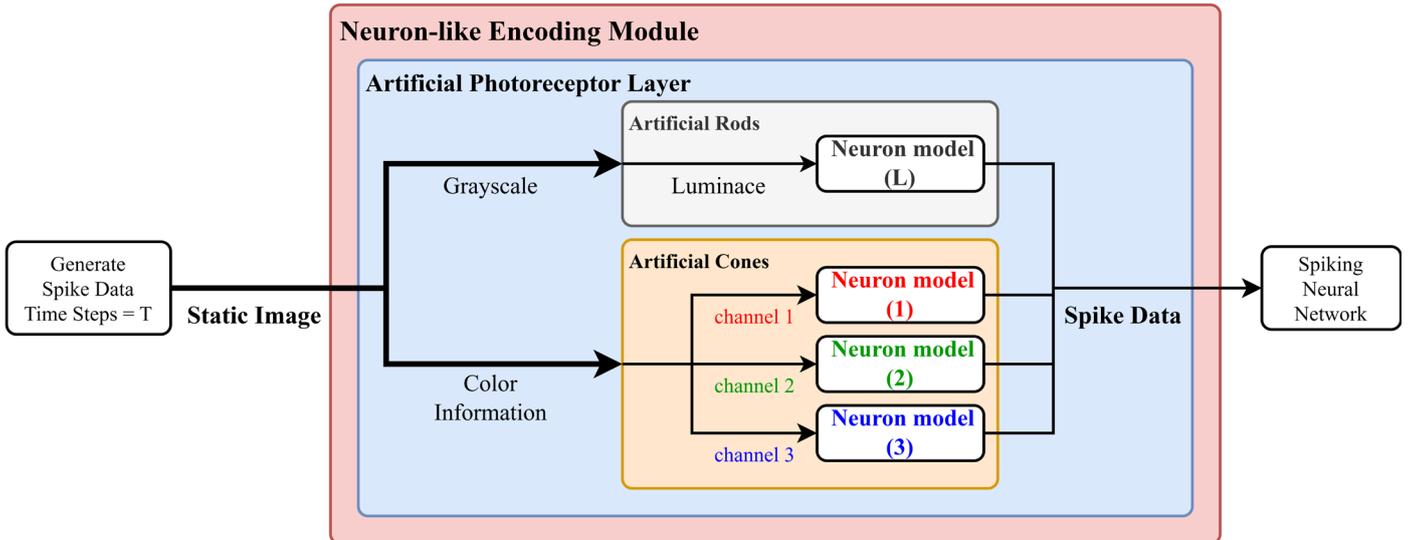

Figure 1. System Architecture Overview. Assuming constant incident light energy and no external noise within the exposure time, a static image can be continuously fed into the Neuron-like Encoding Module to simulate an instantaneous visual signal. We propose converting static visual input into spike-based representations using biologically inspired neuron models. The artificial visual photoreceptor layer, composed of artificial rod and cone cells, processes the visual signal along different dimensions to encode both chromatic and luminance information into spikes, thereby increasing the information density of the spike stream. In this study, all artificial neurons are modeled using the Integrate-and-Fire (IF) neuron model. The number of discrete time steps, denoted as T, can be any positive integer. In our implementation, we set T=256, meaning that each static image is input into the Neuron-like Encoding Module over 256 consecutive time steps. This process produces a discrete spike sequence of temporal length 256, effectively capturing the visual content in a spike-based format suitable for subsequent processing by spiking neural networks.

## II. LITERATURE REVIEW

In this Section, II.A begins with a review of imaging device architectures to provide an understanding of the principles and processes involved in image formation. II.B then revisits the concepts of neuromorphic computing and SNNs, outlining the foundational objectives of neuromorphic computing and summarizing recent advancements in SNN research. In II.C, we examine the existing input data types used in SNNs, exploring the relationship between data characteristics and network performance, as well as how these input modalities have been employed in recent studies.

*A. Imaging Pipeline of Visual Sensors*

In computer vision systems, imaging devices function as the "eyes" of a computer, capable of capturing and converting external light into discrete electronic signals. Modern imaging devices primarily include conventional cameras and event-based cameras, which generate static images and event data, respectively. In this section, we explain the underlying principles of both imaging modalities to understand how visual data is currently acquired.

Conventional cameras remain the primary method for producing color images. During the exposure period, incident light enters the camera through the shutter, passes through internal optical components, and reaches the image sensor [4]. Most modern conventional cameras utilize CMOS (Complementary Metal-Oxide-Semiconductor) sensors, which integrate both NMOS and PMOS transistors into the same circuit unit. NMOS and PMOS refer to N-type and P-type Metal-Oxide-Semiconductor Field-Effect Transistors (MOSFETs), respectively. A MOSFET has four terminals: source, drain, gate, and body. When the substrate is P-type and the source and drain are N-type, the transistor is classified as NMOS; the reverse configuration results in a PMOS. MOSFETs consume power primarily during switching and exhibit low leakage current, making them power-efficient for circuit design. N-type and P-type semiconductors are differentiated by doping characteristics, where N-type contains more free electrons than holes, and P-type the opposite—hence their complementary nature in CMOS design.

In a conventional camera, CMOS sensors themselves are not photosensitive and require photodiodes to convert light into electrical power. Photodiodes generate a photocurrent through the photoelectric effect: incident photons with sufficient energy excite electrons into free carriers, creating electron-hole pairs. This process increases the stored electrical energy in the pixel capacitor.

During image acquisition, the shutter opens for a specified duration, allowing light to enter the camera—a process known as exposure. The light travels through the aperture and dielectric layers before reaching the photodetector array. The aperture controls the amount of light entering the system and affects depth of field. As incident light does not directly carry color information, a Bayer color filter is applied to separate incoming light into red, green, and blue channels. To maximize light collection, a microlens array is positioned above the photodiodes to focus light onto the sensitive regions.

While the shutter remains open, light continuously hits the sensor. The resulting photocurrent charges the capacitor, a process termed integration, and the duration is called the integration time or exposure time. However, the capacitor has a limited storage capacity; once saturated, it cannot accumulate additional energy. After the exposure period, the stored charge is read out as a voltage, amplified, and converted into digital signals by an analog-to-digital converter.

The choice of integration time significantly influences image quality and temporal resolution. Overexposure can result in blown highlights, whereas underexposure may obscure shadow details. Furthermore, longer integration times reduce temporal resolution (i.e., fewer frames per second), whereas shorter integration times allow higher frame rates.

In contrast, event cameras represent a novel class of imaging sensors that detect changes in light intensity rather than capturing full-frame images [5, 6]. Inspired by the human retina and its sensitivity to fast-moving stimuli, event cameras emulate the behavior of rod cells (which are highly sensitive to light), bipolar cells, and ganglion cells that perform early-stage visual filtering. These sensors output asynchronous event data containing four elements: x-position, y-position, timestamp, and polarity—collectively referred to as event data.

An event camera pixel comprises a photodiode, two capacitors, and a comparator, functionally analogous to retinal rod cells, bipolar cells, and ganglion cells, respectively. At each pixel, light energy at two successive time points is stored in separate capacitors. The comparator then evaluates the energy difference. If the newer intensity surpasses the previous one by a threshold, a positive event is triggered (polarity = +1); if it decreases beyond a threshold, a negative event is recorded (polarity = −1). This mechanism produces asynchronous, high-speed data streams.

Although both conventional and event cameras require light to strike the photodetector array for image formation, a key distinction lies in the lack of integration in event cameras. Event-based sensors offer several advantages over conventional cameras. First, they exhibit a high dynamic range, enabling reliable performance under varying illumination conditions. Second, they possess exceptionally high temporal and spatial resolution, eliminating motion blur and increasing effective frame rates. Third, since they only report significant changes in intensity, their power consumption is inherently lower.

However, a major limitation is that event cameras, modeled after rod cells, cannot detect color. Consequently, event data only convey changes in luminance, lacking chromatic and texture information. This restricts their applicability in tasks that require comprehensive scene understanding, particularly in scenarios demanding color discrimination or static context.

*B. Neuromorphic Computing and Spiking Neural Networks*

The pursuit of endowing machines with human-like visual perception has driven researchers to design systems inspired by biological processes, aiming to replicate these mechanisms in artificial systems as faithfully as possible. While significant progress has been made—evident in the development of cameras and ANNs with remarkable performance—there remains a substantial disparity between the high power consumption of machine computation and the low energy efficiency of biological neural signaling. To bridge this gap without compromising performance, it is essential to emulate the detailed operational behavior of biological neurons. This ambition defines the field of neuromorphic computing.

Neuromorphic computing, like DL, was proposed several decades ago [1]. However, early efforts were hindered by limitations in hardware and mathematical modeling. The resurgence of DL and the advent of new computational methods have revitalized interest in this area. The central premise of neuromorphic computing is to enable information transmission and processing through spike-based signaling, mimicking the dynamics of the human nervous system. For machines, spike signals—represented as binary values (0 and 1)—offer advantages such as reduced storage requirements, faster computation, and lower power consumption, making them an ideal data type.

In biological neurons, dendrites receive incoming spike signals, causing sodium ions (Na$^+$) to enter the cell and increase the membrane potential. When this potential reaches a threshold, the neuron fires an action potential through its axon, releasing potassium ions (K$^+$) to return to a resting state. Notably, neurons are only active during spike transmission and remain at a resting potential of approximately -70 mV otherwise [1].

The earliest model to describe spiking neurons was the Hodgkin-Huxley model, which used mathematical s and electrical circuits to capture the dynamics of real neurons. Due to its complexity, it was later simplified to the Leaky Integrate-and-Fire (LIF) model, as described by (1). The LIF model omits sodium and potassium ion channels, modeling the neuron as a simple RC circuit. Incoming spikes charge the capacitor, mimicking the accumulation of membrane potential, which naturally decays over time:

$$U_{mem}(t) = I_{in}(t)R + [U_0 - I_{in}(t)R]e^{-\frac{t}{RC}} \qquad (1)$$

Here, $U_{mem}(t)$ denotes the membrane potential at time $t$, $I_{in}(t)$ is the input current, $R$ is the resistance, $U_0$ is the initial potential, and the exponential term represents the natural decay over time. When the accumulated potential reaches a predefined threshold, the capacitor discharges and generates a spike, which is transmitted to the next neuron [1, 7, 8].

Artificial neural networks based on the LIF model are referred to as SNNs. Unlike traditional ANNs, SNNs cannot directly apply gradient descent due to the non-differentiable nature of spikes—a problem known as the Dead Neuron Problem [8, 9]. To address this, one approach involves converting pretrained CNNs into SNNs using ANN-to-SNN conversion, where layer weights are transferred and integrated with spiking modules. Although this method can produce performance comparable to CNNs and Transformers, it typically requires a large number of time steps, resulting in prolonged inference times. The ANN-to-SNN approach is typically implemented following the procedure illustrated in Fig. 2(a). First, an ANN is trained using static images. The trained ANN is then converted into an SNN architecture, after which the images undergo spike encoding before being processed by the SNN. An example of such applications can be found in [10].

An alternative line of research focuses on training SNNs directly. This involves approximating spike functions with steep, differentiable curves during backpropagation. Methods such as SuperSpike [11] and SLAYER [12] were among the earliest solutions designed to tackle the Dead Neuron Problem. SLAYER (Spike Layer Error Reassignment) employs a specially designed loss function to distribute temporal error across time steps and uses surrogate gradients to enable weight updates, allowing the network to learn complex spatiotemporal patterns.

While methods like SLAYER are effective, their complexity often limits their portability across other SNN architectures. To overcome this, the Surrogate Gradient Descent(SGD) approach was proposed [13], providing a general and adaptable solution for direct training of SNNs. One of the earliest and most notable implementations of this method is DECOLLE (Deep Continuous Local Learning) [14]. DECOLLE uses custom spiking neurons—DECOLLE Neurons—with four internal variables (P, Q, R, S) to perform feature extraction, firing, and reset operations in conjunction with convolutional layers. Its core innovation lies in a local learning strategy, where each layer contains its own readout module and corresponding loss function. These layer-wise losses are aggregated and optimized using surrogate gradients, allowing efficient training while minimizing memory usage and decoupling performance from time step constraints. However, this benefit comes at the cost of increased training time as the network depth grows.

The success of SGD methods has prompted the incorporation of CNN and Transformer architectures into SNNs. Examples include S-ResNet [15], SEW-ResNet [2], and SpikFormer [3], each aiming to push the performance of SNNs to match or surpass that of conventional architectures. S-Res-Net (Spiking ResNet) [15] was the first to introduce residual connections into SNNs, though it suffered from instability during training, such as gradient vanishing and explosion. SEW-ResNet (Spike-Element-Wise ResNet) [2] addressed these issues by redesigning the residual blocks to operate element-wise on spike data, enabling stable and scalable training of deep SNNs, and marking the first successful training of SNNs exceeding 100 layers.

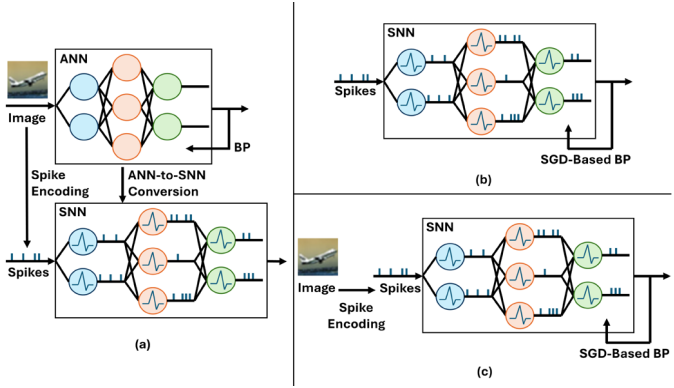

Figure 2. Schematic illustration of training workflows under different approaches. (a) ANN-to-SNN conversion, (b) surrogate gradient descent with spike-based data, and (c) surrogate gradient descent with non-spike-based data.

Building on these innovations, Transformer architectures have also been adapted for use in SNNs. SpikFormer [3] is the first SNN to incorporate Transformer structures by introducing Spiking Self-Attention (SSA). SSA converts the query, key, and value operations from floating-point representations to spike-based formats, implementing self-attention via simple addition operations. This significantly reduces computational complexity and power consumption, making it compatible with the low-power philosophy of SNNs while leveraging the powerful representational capacity of Transformers.

For approaches that train SNNs directly using SGD, the procedures depicted in Fig. 2(b) and Fig. 2(c) are typically followed. If the input data are already in spike form, the training process adheres to the workflow shown in Fig. 2(b). Conversely, if the input data are not spike-based, the workflow in Fig. 2(c) is adopted, wherein the data must first undergo spike encoding before being processed by the SNN. An real example of such applications can be found in [16].

### C. Input Data and Acquisition Methods for Spiking Neural Networks

Recent studies [2, 3, 12, 14, 17-33] have demonstrated the diversity of input data used in SNNs. Regardless of the data source, input must be transformed into spike-based representations through a Spike Encoding Layer (SEL) to comply with the neuromorphic computing paradigm and enable SNNs to process spatiotemporal spike signals. To date, SELs have been realized using various techniques, including classical neural coding schemes, simulated event-based data, and dedicated SNN Encoding Layers (ELs).

Earlier works, adhering to the principles of neuromorphic computing, primarily focused on using spike-based inputs, often derived by converting existing data using biologically inspired neural coding mechanisms. These neural coding techniques originate from neuroscience, where researchers analyze the frequency and timing of neuronal spikes to interpret the information they carry. The two dominant encoding schemes are Rate Coding and Temporal Coding.

Rate Coding encodes information by quantifying the number of spikes emitted by a neuron over a given period, yielding the Mean Firing Rate. Three primary variants exist:

1. Count Rate Code calculates the firing rate of a single neuron by counting the number of spikes within a time window $T$, as defined in (2):

$$\nu = N_{Spike}(T) \quad (2)$$

where $\nu$ denotes the mean firing rate, $N$ Spike is the number of spikes within time $T$.

2. Density Rate Code determines the average firing rate across multiple trials $K$, using a short interval $\Delta T$, as shown in (3):

$$\rho = \frac{1}{\Delta T} \cdot \frac{N_K(t:t+\Delta T)}{K} \quad (3)$$

3. Population Rate Code calculates the average spike activity across a population of n neurons at a specific time window, as shown in (4):

$$A = \frac{1}{\Delta T} \cdot \frac{N_{act}(t:t+\Delta T)}{n} \quad (4)$$

Here, $N_{act}$ refers to the total spike count among all neurons during the interval.

In contrast, Temporal Coding encodes information based on the precise timing of the first spike in response to a stimulus—commonly referred to as Time-to-First-Spike [1, 7, 8, 34, 35].

Based on these concepts, image-to-spike conversion techniques have been developed:
- Rate Coding maps pixel intensities to the firing probability of spikes. Brighter pixels result in higher spike rates. The resulting spike trains are often modeled as Poisson Spike Trains.
- Temporal Coding encodes pixel values as spike latencies—brighter pixels spike earlier.

Both methods are pixel-value-driven. Rate Coding introduces stochasticity by encoding frequency-based information, and has been adopted in some recent SNN studies. Temporal Coding, on the other hand, emphasizes precise spike timing. A simplified schematic is presented in Fig. 3(a).

Beyond Rate and Temporal Coding, other spike encoding strategies include Phase Coding and Burst Coding:
- Phase Coding converts pixel intensity into phase offsets relative to a background periodic signal, leveraging phase differences to transmit information.
- Burst Coding produces a burst of spikes within a short time window, with both the inter-spike interval and number of spikes reflecting pixel intensity. A classical method is Inter-Spike Interval (ISI) Coding.

Despite their potential advantages in reliability, representational capacity, and reduced latency, Phase Coding and Burst Coding face practical challenges when applied to image encoding. For instance, the variability in cycle frequency and period in Phase Coding complicates the spike representation. Similarly, defining short durations and inter-spike intervals in discrete systems for Burst Coding is nontrivial [34, 35]

Spike data generated through Rate or Temporal Coding is commonly referred to as Poisson Spikes. These were widely used in early SNN studies to train and evaluate networks.

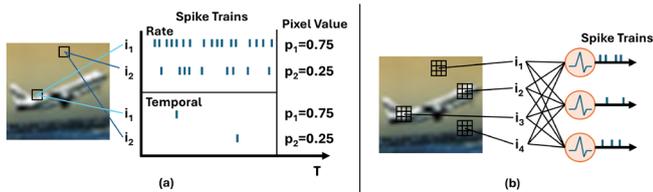

Figure 3. Schematic illustration of encoding methods. (a) neural encoding and (b) encoding layer.

While such representations helped investigate the computational properties of artificial spiking neurons, they often yielded suboptimal performance. This limitation arises because neural coding schemes like Rate or Temporal Coding were originally designed to interpret biological spikes, not to encode complex information-rich signals such as images.

Moreover, these biological coding methods typically focus on a single spike attribute. Rate Coding emphasizes frequency, while Temporal Coding emphasizes timing—resulting in only partial representations. For example, spikes generated through Rate Coding lack temporal resolution, whereas those from Temporal Coding lack frequency-based richness.

In early-stage research—when SNN training techniques were still in their infancy—many studies relied on synthetic or grayscale stimuli, such as sine waves or intensity maps, to generate stable Poisson Spikes. These spike trains were then used to validate the functionality of newly proposed SNN architectures by observing the network's responses and verifying basic behaviors.

Recent studies [2, 3, 12, 14, 17-20, 22-31] have widely adopted event data for training SNNs and evaluating their performance. Among the commonly used datasets are N-MNIST [36], CIFAR10-DVS [37], and DVS128-Gesture [38], all of which were recorded using event-based cameras. Specifically, N-MNIST and CIFAR10-DVS were generated by presenting static datasets—MNIST [39] and CIFAR10 [40], respectively—to an event camera to simulate dynamic stimuli. For N-MNIST, the event camera was mounted on a differential motion platform that oscillated to induce apparent motion from static MNIST images. In the case of CIFAR10-DVS, the event camera remained stationary in front of a screen displaying CIFAR10 images, which were moved across the screen in a controlled manner to generate temporal luminance changes captured as event data.

In contrast, DVS128-Gesture differs significantly in nature. While N-MNIST and CIFAR10-DVS are synthetic dynamic datasets derived from originally static content, DVS128-Gesture is a genuinely dynamic dataset, collected by recording real human hand gestures (e.g., waving) using an event camera.

Due to the high cost and limited accessibility of event-based cameras compared to conventional frame-based imaging devices, several recent studies have proposed event data emulation techniques to simulate event streams from RGB videos. Notable approaches include Vision Sensor Behavioral Emulator (VSBE) [41], Pixel-to-NVS (PIX2NVS) [42], ESIM [43], and v2e [44]. These methods typically perform logarithmic luminance transformations on pixel values to enhance brightness sensitivity and, based on the principles underlying event camera operation, compute inter-frame luminance differences over time to generate synthetic frame-based event data. Through such simulations, traditional RGB videos captured with conventional cameras can be transformed into event streams.

However, due to the design limitations inherent to both real and simulated event cameras, neither real-world nor emulated event data contains color information. These data represent only changes in brightness occurring during motion, thereby omitting chromatic content entirely.

Since the SGD method [13] successfully addressed the Dead Neuron Problem, SNNs have rapidly integrated techniques from deep learning (DL), leading to significant maturation of the field. Many objectives that were previously unattainable have now become feasible, transforming SNNs from systems limited to observing input-output variations into viable next-generation AI architectures.

In the study by Lei et al. [30], prior methodologies were summarized, and four models were evaluated using the CIFAR10 dataset: (1) an ANN trained with static images, (2) an SNN converted from ANN using the ANN-to-SNN method, (3) an SNN trained with Poisson spikes generated via rate coding, and (4) an SNN trained directly using static color images. The latter two SNNs—i.e., those generated via ANN-to-SNN conversion and those trained with Poisson spikes—were tested using Poisson spike representations derived from rate-coded image data. In contrast, both the ANN and SNN trained directly on static color images were tested with standard static input. The results showed that the ANN model achieved the highest accuracy, followed by the converted SNN and the directly trained SNN. The lowest performance was observed in the SNN trained using rate-coded Poisson spikes.

Notably, since static color images are not inherently spike-based, the aforementioned study introduced an EL at the input stage of the SNN to ensure spike-based information transmission. A simplified schematic is presented in Fig. 3(b). This EL consists of a convolutional layer for preliminary feature extraction, followed by a firing module to convert the extracted features into spike signals. This approach may partially explain the improved performance of SNNs trained directly on static images, as the convolutional preprocessing could help produce meaningful spike patterns that are easier for the network to learn.

To further optimize SNN performance, recent studies [2, 3, 17-27] have increasingly adopted static color images as training input and incorporated EL-based mechanisms to perform both feature extraction and spike conversion. Static images, widely used in computer vision tasks, are typically acquired using conventional cameras and stored in 8-bit unsigned integer (uint8) format, with pixel values ranging from 0 to 255. Grayscale images, which store only brightness information, use a single channel, whereas color images use three channels representing visible red, green, and blue light intensity, respectively.

Unlike spike-based data, color static images preserve color information across three channels and represent light intensity using discrete integer values, thus maximizing data fidelity. However, as with conventional frame-based capture, static images lack any inherent temporal relationship. Each frame in a video is an independent sample, unaffected by preceding or subsequent frames. As a result, temporal continuity is absent, making it impossible to predict the next frame from a single static image.

To fulfill the temporal learning requirements of SNNs, recent research has proposed repeatedly inputting the same static image over multiple timesteps, allowing membrane potentials and spike dynamics to function as intended and enabling the network to express its temporal behavior effectively.

It is important to note that EL-based architectures are fundamentally frame-based systems, yet they are capable of handling both static and dynamic datasets. When used with event-based data, these models convert event streams into event frames through temporal framing techniques. Consequently, the input data in such architectures are not true spike trains but are instead processed via EL to perform convolution and spike conversion.

Recent studies have proposed two main approaches for using static color images with SNNs. The first applies standard normalization, adjusting pixel values to have zero mean and unit variance. This process introduces negative pixel values, mimicking the positive and negative polarity inherent in event-based data [18, 20-23, 29, 32]. The second approach, consistent with conventional CNN training practices, rescales pixel values to the [0, 1] range before feeding them into the model [2, 3, 17, 19, 24-27]. Regardless of preprocessing strategy, both approaches signify a shift away from purely spike-based input formats. Nonetheless, this shift has led to substantial improvements in SNN performance—often approaching that of CNNs—and has enabled the integration of color information into SNN architectures.

While this direction of research has successfully maximized the performance of contemporary SNNs and broadened their applicability, it also diverges from the original bio-inspired ideal of SNNs operating solely on spike data. We consider this a domain-specific trade-off, reflecting a pragmatic balance between theoretical fidelity and empirical performance.

We posit that the observed differences in performance between SNNs trained with Poisson spike trains and those trained directly with static color images can be attributed to distinct underlying factors. For SNNs trained using Poisson spikes, although prior studies often applied rate coding independently to each of the three color channels—thereby preserving the color information as much as possible—the intrinsic limitation of rate coding lies in its unidimensional encoding of frequency information, without capturing temporal dependencies. As previously discussed, the generation of Poisson spikes is modeled as a series of probabilistically independent events, which prevents the network from learning meaningful spike timing relationships. This independence significantly limits the performance of rate-coded SNNs.

In contrast, SNNs trained directly with static color images benefit from the rich informational content encoded in the input data. Compared to Poisson spike trains, static color images inherently preserve higher-order spatial and chromatic features, making it unsurprising that they yield better performance. However, despite the performance advantage of combining ELs with static color images, the biological plausibility of such an approach remains questionable. It is thus necessary to examine whether this method diverges from the structure and mechanisms of the biological visual system.

According to previous literature [45, 46], the human visual system comprises three primary components: the eyeball, the optic nerve, and the brain. The eyeball captures external light stimuli and converts them into bioelectrical and spike signals. These signals are then transmitted via the optic nerve, which performs preliminary preprocessing, before being relayed to the brain for further integration and interpretation—enabling perception and appropriate behavioral responses.

The eyeball receives incoming light through the pupil, which is focused by the lens onto the retina to generate clear biological visual signals. The retina is densely packed with photoreceptor cells, especially in the fovea, the central region responsible for sharp vision. To focus on an object of interest, humans naturally adjust their gaze to ensure that reflected light converges precisely at the fovea. Adjacent to the fovea lies the optic disc, a physiological blind spot where no photoreceptors exist, as it serves as the exit point for signals traveling through the optic nerve.

There are two main types of photoreceptors on the retina: rod cells and cone cells. Rod cells, numbering approximately 125 million, are highly sensitive to light and are distributed broadly across the retina. They can fire spikes even under very low illumination and are thus essential for scotopic (low-light) vision. However, rod cells respond slowly and lack the spatial acuity and color sensitivity necessary for high-resolution perception. As a result, under dim conditions, humans often perceive motion without being able to discern fine details or color.

Conversely, cone cells, numbering approximately 6–7 million, are concentrated in the fovea and decrease toward the periphery. They require well-lit conditions to function and are nearly inactive in darkness. Each cone cell contains one of three types of opsins sensitive to different wavelengths of light (564 nm, 534 nm, and 420 nm), allowing for trichromatic color vision through differential firing patterns. Consequently, the spikes generated by cone cells carry color-specific information, enabling the brain to construct a full-color perceptual experience.

The optic nerve, located at the rear of the eye, transmits the spike signals from the retina to the brain while also performing initial signal processing. It comprises neurons oriented both horizontally and vertically. Horizontal cells propagate signals laterally to neighboring neurons, modulating their activity through excitatory or inhibitory interactions to facilitate local contrast and context processing. Vertical neurons, including bipolar cells and ganglion cells, relay signals toward the brain. Bipolar cells, situated immediately downstream of the photoreceptors, exist in ON and OFF varieties and are responsible for transmitting spike activity. Ganglion cells, categorized into P-type and M-type, exhibit different response characteristics: P-type ganglion cells (about 80% of all ganglion cells) transmit fewer signals at slower rates, with higher sensitivity to color but lower sensitivity to motion. In contrast, M-type ganglion cells transmit more signals rapidly, respond strongly to motion and luminance changes, but are relatively insensitive to color.

Following initial processing, the spike trains are projected to the primary visual cortex, where more complex feature extraction occurs before further integration in the prefrontal cortex, culminating in perceptual recognition and behavioral response.

From this biologically grounded perspective, we identify two major inconsistencies in the use of static color images and ELs for training SNNs. First, regarding repeated input of static images: in real biological vision, even under continuous observation of a still object, the images perceived by the retina are never numerically identical across frames. This is due to inherent sensor noise, temporal fluctuations, environmental dynamics, and quantization artifacts introduced during the analog-to-digital conversion (e.g., mapping light intensity to 8-bit integers). Therefore, the practice of feeding identical static frames multiple times into SNNs does not reflect biological reality.

Second, with respect to the function of ELs, biological neurons cannot simultaneously perform multiple distinct functions. In the visual system, neuronal specialization is a fundamental principle: a neuron responsible for signal transduction (e.g., photoreceptors) does not also perform complex processing or encoding, which is delegated to downstream neural structures (e.g., bipolar and ganglion cells). Thus, the ELs in artificial

models—designed to both extract features and convert them into spikes—violate this division of labor seen in biological systems.

In conclusion, we argue that static color images, as non-spike-based inputs, should not be used directly to train SNNs via repeated presentation. This approach introduces discrepancies with the operational principles of biological vision and risks undermining the neuro-inspired authenticity of SNN design.

## III. METHOD

In this Section, Section A introduces a biologically inspired neural spike encoding method based on the human visual and nervous systems. Section B presents the design of an artificial visual photoreceptive layer aimed at closely emulating the functionality of the retina. Finally, Section C outlines the overall system framework proposed in this study.

*A.  Neuron-like Encoding*

As previously discussed, recent SNN research frequently employs static images as input data, where an EL is used to convert the images into spike signals. However, from the perspective of neuromorphic computing, it is expected that only spike-based signals are processed throughout the system. Therefore, it is biologically implausible for static images to undergo convolutional operations in the EL before being transformed into spikes. Furthermore, convolving non-spiking data introduces an additional layer of biological inconsistency. On the other hand, applying rate or temporal coding to convert static images into Poisson spikes results in spike trains that carry only partial information. Although event data aligns well with the neuromorphic computing paradigm, it still suffers from incomplete visual representation.

To simultaneously address the limitations of Poisson spikes—which convey only partial information—and the issue of non-spike inputs, we reconsider the fundamental process by which spikes should be generated, drawing inspiration from the biological nervous system. Fundamentally, all spike signals in the nervous system are generated by neurons and are transmitted from the peripheral to the central nervous system. This raises a critical question: what type of input is received by the neurons at the system's periphery? If all spike signals originate from neurons, then the earliest neurons in the processing chain must also receive spike signals. However, these initial neurons have no upstream neurons and thus cannot receive spike-based inputs. This suggests that the input to the first neurons must originate from external stimuli, which are then converted into spikes.

In the human visual system, the sensory receptors—namely, rod and cone cells located in the retina—are responsible for converting incoming light into spike signals. Importantly, these sensory neurons do not interact or exchange information with each other and lack the ability to perform real-time information processing. Signal integration and information exchange only occur in subsequent neurons, such as bipolar cells, horizontal cells, and ganglion cells. Therefore, we argue that to generate spike signals that simultaneously preserve both rate and temporal coding characteristics, such signals must be generated by neurons. Moreover, these converting neurons should not interfere with one another or process external information jointly.

Among various neuron models in computational neuroscience, the LIF model is one of the most well-established and commonly used. It captures essential biological properties, including the decay of membrane potential over time. However, in discrete-time systems with relatively low temporal resolution, the decay mechanism can lead to excessively long time steps and increased computational cost. Therefore, we adopt the Integrate-and-Fire (IF) model, a simplified idealization of the LIF model that ignores membrane potential decay. The IF model allows for stable and rapid spike generation over short time spans, making it suitable for spike visualization and verification purposes. Consequently, the IF neuron is selected as the principal model for this study.

We hypothesize that spike signals generated by neuron models such as the IF neuron will inherently exhibit both temporal and frequency-related characteristics. These spike trains are expected to exhibit both rate and temporal coding behaviors when visualized, and may outperform traditional spike-encoded data in SNN-based classification tasks. While we do not claim that spike signals generated by the IF neuron will necessarily outperform EL-based systems—especially given the idealized nature of the IF model—we note that EL introduces biologically unrealistic assumptions despite its proven efficacy in enhancing SNN performance through convolutional preprocessing.

We refer to the use of neuron models to convert external stimuli into spike signals as "neuron-based spike encoding." Depending on the specific model employed, this process may be referred to as LIF coding (using the LIF model) or IF coding (using the IF model). The IF model accumulates input energy as (5):

$$E_{mem}(T_{step}) = E_{mem}(T_{step} - 1) + E_L(T_{step}) \quad (5)$$

where $E_{mem}(T_{step})$ denotes the membrane potential at time step $T_{step}$, and $E_L(T_{step})$ represents the external light energy received at the same time step. A spike is generated if the membrane potential exceeds the firing threshold $\theta$, as defined in (6):

$$Spike(T_{step}) = \begin{cases} 1, E_{membrane}(T_{step}) > \theta \\ 0, elsewise \end{cases} \quad (6)$$

When a spike is triggered, the membrane potential is reduced by the threshold energy, as described in (7):

$$E_{mem}(T_{step}) = E_{mem}(T_{step}) - Spike(T_{step}) \times \theta \quad (7)$$

In practical systems, traditional cameras are typically used to capture visual data. Thus, the input stimulus can be reasonably modeled as static images. Camera-based image formation requires exposure time to accumulate light energy in capacitors, which is then converted into pixel values. This process introduces minor fluctuations due to thermal and electronic noise. The accumulation of pixel intensity over the exposure period can be expressed as:

$$Pixel\ Intensity(T_{Exposure}) = \frac{\int [E_L(T) + E_N(T)]}{T_{Exposure}} \quad (8)$$

Under idealized assumptions, where noise $E_N$ is negligible, and light intensity remains constant over time, (8) simplifies to (9):

$$Pixel\ Intensity(T_{Exposure}) = E_{L\_Avg} = \frac{\int E_L(T)}{T_{Exposure}} \quad (9)$$

Assuming an 8-bit unsigned integer image format, pixel values range from 0 to 255. These are commonly normalized to the range [0, 1] as shown in (10):

$$0 \leq \frac{Pixel\ Intensity(T_{Exposure})}{MAX_I} = E_{L\_normalized} \leq 1 \quad (10)$$

where MAX_I = 255. Based on s (9) and (10), we discretize the exposure time $T_{Exposure}$ into 256 sampling steps, each representing a fixed light intensity value. Assuming a firing threshold $\theta = 1$ and initial membrane potential of 0, no spike is generated at the first step. Spiking begins from the second time step and continues until step 256, where the membrane potential resets to zero. If extended beyond 256 steps, the spiking pattern will repeat cyclically.

For example, if the pixel value is 1, its normalized light intensity becomes 1/255. Over 255 time steps, the membrane potential reaches 1. At the 256th step, a spike is triggered, and the potential resets to 1/255. This cycle repeats every 255 steps. Such behavior is depicted in Fig. 4(a), where inputs of 0, 0.25, 0.5, 0.75, and 1 correspond to pixel values of 0 through 4, producing spikes every 4 steps as expected.

Therefore, under an 8-bit format and standard normalization, the IF model can generate spike trains over 256 discrete time steps that faithfully encode the entire content of a static image.

### B. Artificial Photoreceptor Layer

It is well-established that the human retina contains two primary types of sensory neurons: cone cells and rod cells. The spike signals generated by these two types of neurons jointly construct human visual perception; the absence of either form of information would inevitably compromise visual experience. As discussed in Section II.C, static color images currently serve as the most comprehensive format for preserving incident light energy, and they most closely approximate human visual perception. These images preserve both chromatic (color) and luminance (brightness) information to a near-maximum extent. In order to maintain the integrity of external visual stimuli, we propose that both chromatic and luminance information should be converted into spike-based signals in a biologically inspired manner and simultaneously provided to SNNs during training. This design is intended to ensure that SNNs can learn from inputs containing rich and complete visual information. Accordingly, we introduce the concept of an Artificial Visual Photoreceptor Layer which consists of Artificial Cone Cells and Artificial Rod Cells.

Inspired by cone cells, and as described in Section II.C, we know that chromatic spike signals are generated not by a single photoreceptor but rather by distinct cone cells that contain different photopigments, each responsive to specific wavelengths of light. This mechanism is functionally analogous to the Bayer filter in traditional digital cameras, which separates incoming light into red, green, and blue channels. Therefore, building upon the neuron-based spike encoding scheme introduced in Section III.A, we construct Artificial Cone Cells by applying a neuron model (e.g., IF neuron) to encode color images directly. The neuron model independently encodes each color channel (Red, Green, and Blue) to produce color spike signals as shown in (11):

$$Spike_C = Encoding(Pixel\ Intensity_c), C = R, G, B \quad (11)$$

Here, Encoding refers to the method used to convert pixel intensities into spike signals. This could be any spike encoding strategy or neuron-based encoding model; in this work, we adopt IF Coding as the default method. C denotes the color channel, with R, G, and B referring to Red, Green, and Blue, respectively. $Spike_C$ represents the spike signal generated by encoding the corresponding pixel intensity of channel C.

Next, inspired by the function of rod cells—as outlined in Section II.C—we note that these cells are highly sensitive to light intensity but are incapable of encoding color. To simulate this functionality, we derive a grayscale image from the original color input, treating it as the global luminance of the scene (i.e., unfiltered incoming light prior to chromatic separation). This grayscale representation is then fed into the Artificial Rod Cells for spike encoding. In doing so, we aim to simulate the complementary roles of retinal neurons in providing complete visual information. This extension leads to the revised form of (11), denoted as (12):

$$Spike_C = Encoding(Pixel\ Intensity_c), C = R, G, B, L \quad (12)$$

Here, L denotes the luminance (or brightness) channel. As a result, the system generates spike signals for each of the Red, Green, Blue, and Luminance channels, collectively representing a more biologically faithful and information-rich visual input for SNN training.

### IV. EXPERIMENT

To evaluate the effectiveness of IF Coding in improving the performance of SNNs, as well as to investigate the impact of temporal resolution on static data when used in conjunction with SNNs, we conducted a series of experiments involving spike visualization, analysis of static images under varying temporal resolutions, and direct training of SNNs with these inputs. In this study, four SNN models were employed to assist with analysis and validation: SLAYER [12], DECOLLE [14], SEW-ResNet [2], and SpikFormer [3], all of which were introduced in Section II.

All experiments were conducted using the CIFAR-10 dataset [40], which contains 60,000 color images of 32×32 pixels, categorized into 10 classes (airplane, automobile, bird, cat, deer, dog, frog, horse, ship, and truck). The dataset is split into 50,000 training images and 10,000 test images. Although traditional CNN-based architectures have achieved excellent performance on CIFAR-10 [47-49], our study focuses solely on the performance variations in SNNs due to different input encoding strategies.

Table A summarizes the model architectures and training environments. Notably, no learning rate scheduler was employed during training. For SLAYER, the original model architecture from its paper was used. For DECOLLE, the number of channels in the hidden layers was adjusted. The SEW-ResNet model was adapted from its original version designed for CIFAR10-DVS, and SpikFormer was implemented following its original CIFAR10-DVS codebase, but reconfigured with parameters suitable for the static CIFAR-10 dataset. For software environments, we recreated the setups described in each model's original publication.

To compare the characteristics of spike-encoded inputs, we first visualized the spike outputs generated by rate coding, IF coding, temporal coding. These spike signals were then used to train the SNNs under consistent conditions, allowing us to evaluate the impact of different encoding strategies on model accuracy. Additionally, we compared these results against models employing EL, as reported in recent literature. We also examined the performance of IF Coding across various color space transformations (LMS, LAB, and YUV), and measured the computational cost, including FLOPs, parameter counts, inference time, and energy consumption, to evaluate differences in efficiency between static and spike-based data formats.

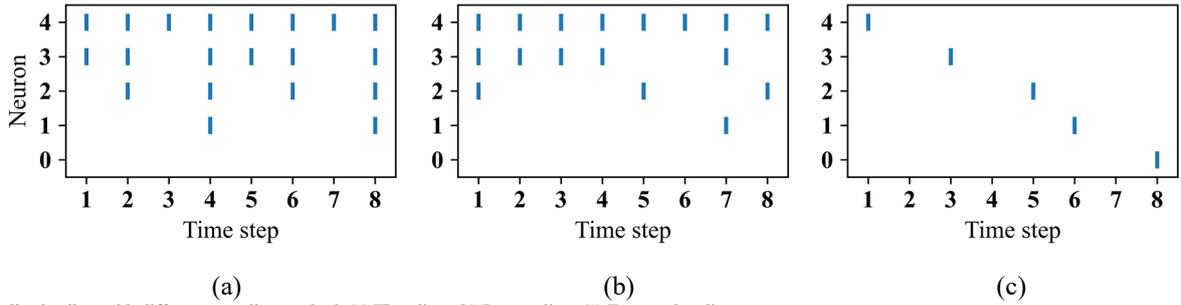

Figure 4. Visualized spikes with different encoding method. (a) IF coding. (b) Rate coding. (c) Temporal coding.

As outlined in Section III.A, we hypothesize that IF Coding represents a promising spike encoding approach. To validate this, we conducted a comparative visualization of spike patterns generated by IF Coding, Rate Coding, and Temporal Coding, assuming five normalized pixel values of 0, 0.25, 0.5, 0.75, and 1. Fig. 4 shows the corresponding spike outputs. In Fig. 4(b), although Rate Coding produces numerous spikes, their probabilistic generation leads to a lack of temporal correlation, making value reconstruction imprecise—for instance, Neuron 1, 2, and 3 reconstruct values of 0.125, 0.375 and 0.625, respectively. Fig. 4(c) illustrates that spikes from Temporal Coding convey temporal information based on signal intensity but result in very sparse spike outputs. In contrast, as shown in Fig. 4(a), spikes generated by IF Coding exhibit both accurate value reconstruction—Neuron 1 yields 0.25, Neuron 2 yields 0.5, and Neuron 3 yields 0.75—and intensity-proportional timing (e.g., first spikes appear at time steps 4, 2, and 1 for Neurons 1–3, respectively). This confirms that IF Coding integrates both rate-based and temporal-based features, thereby validating the efficacy of neuron-based encoding.

To further validate the information richness of spikes generated via IF Coding relative to Rate and Temporal Coding, we trained SNNs using spike inputs derived from grayscale versions of CIFAR-10 images. The grayscale images were independently encoded into spike trains using each of the three coding schemes, with all models configured to operate at 256 time steps to ensure a fair comparison aligned with our encoding assumptions. The resulting performances were evaluated to assess the practical strengths and weaknesses of each encoding method.

We first conducted a comparison of encoding methods under the condition of luminance-only inputs. The results, shown in Fig. 5(a), indicate that across all four SNN models, IF Coding achieves higher accuracy than both Rate Coding and Temporal Coding. While Rate and Temporal Coding yield similar performance in the SLAYER model, Temporal Coding shows a clear advantage over Rate Coding in the other three models. Despite variations due to architectural differences, these findings strongly support the hypothesis that IF Coding provides superior spike representations, as initially anticipated.

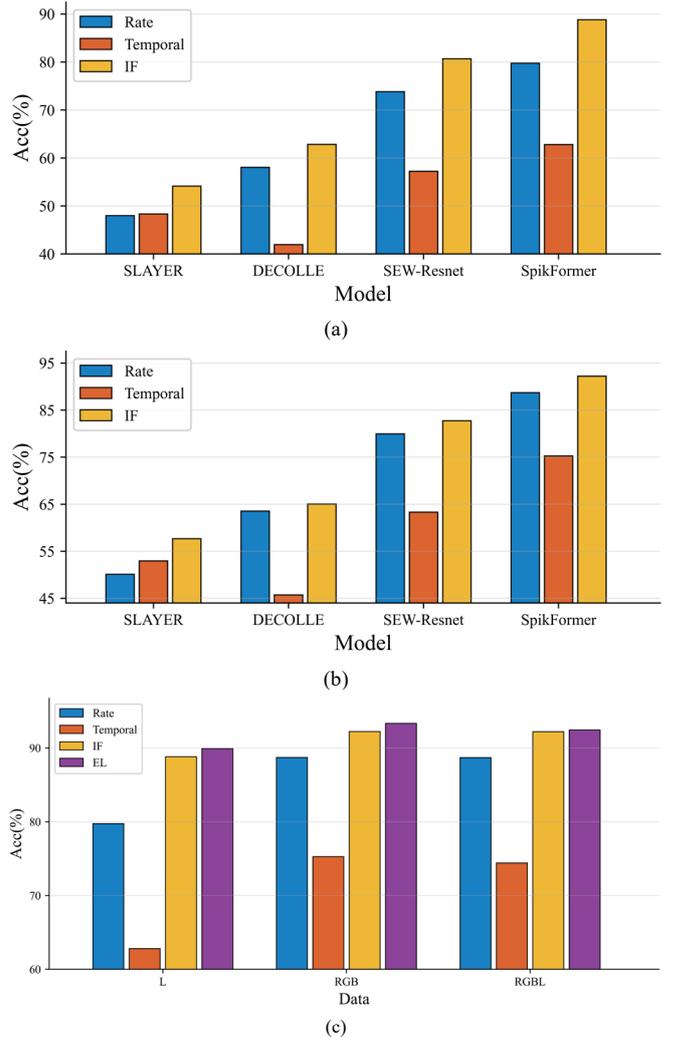

Figure 5. Accuracy of SNNs using different encoding methods. (a) Trained with luminance information (b) Trained with color information. (c) SpikFormer trained with different coding and data combinations.

Subsequently, we explored the encoding method comparison under the more common CNN-like setup, using only color information as input. As shown in Fig. 5(b), IF Coding again achieves the highest accuracy among the three encoding schemes. The observed trends are consistent with those from the luminance-only setting, suggesting that increasing the effective dimensionality of spike-encoded inputs can indeed enhance SNN performance, thereby further validating the proposed IF Coding approach.

We further examined the performance of the SpikFormer model under three input conditions: using luminance information only, using color information only, and using both luminance and color information. As shown in Fig. 5(c), a clear improvement in classification accuracy is observed when replacing or augmenting the input data, with the best performance achieved using color information. Furthermore, among the different encoding schemes, IF coding consistently yielded the excellent results, in line with our initial hypothesis, although its performance remained slightly lower than that of the EL method.

However, when both grayscale and color information were combined, the resulting effect on accuracy was marginal. Notably, in the case of IF Coding, the performance slightly declined, as shown in Fig. 6(a). We propose two possible explanations for this phenomenon. First, from a mathematical perspective, grayscale images can be derived from color images using predefined conversion formulas. While such transformations are not strictly reversible, they imply a unidirectional dependency between the two representations. Second, from a biological standpoint, rod cells are primarily responsible for scotopic (low-light) vision. Under well-lit conditions, their contribution may be minimal or inactive. Therefore, during signal integration in the neural layers, spike inputs from rods might offer little additional information or even introduce unnecessary redundancy, leading to diminished performance.

To thoroughly evaluate whether IF Coding still exhibits any limitations in information representation, we conducted a comparative analysis against the EL approach, which directly converts static images into spike sequences. This comparison was carried out exclusively on models that were originally designed with EL, namely SEW-ResNet and SpikFormer. As illustrated in Fig. 6(b), the accuracies achieved using IF Coding and EL are nearly equivalent in the convolutional architecture of SEW-ResNet. However, in the Transformer-based SpikFormer, EL consistently outperforms IF Coding. Further examination of the validation accuracy curves during training, as shown in Fig. 7, reveals that EL maintains a slight yet consistent advantage over IF Coding throughout the later training stages. These results suggest that while IF Coding approximates the informational completeness of static RGB inputs, a discernible gap remains—particularly in more complex architectures such as SpikFormer.

While the observed differences between IF Coding and EL align with our expectations, we propose three main explanations. First, the IF model is an idealized neuron model that omits biological elements such as membrane potential decay and ionic channels, potentially limiting the fidelity of spike encoding. Second, the EL performs a convolutional feature extraction step prior to spike generation, ensuring that only salient features from the static image are encoded into spikes, thereby facilitating more efficient learning. Lastly, from a temporal resolution perspective, IF Coding encodes the image across 256 discrete time steps, with each frame representing a partial view of the image. In contrast, EL methods can present the entire image at each time step, albeit over fewer steps. This means that when time steps are aligned, each EL frame effectively encompasses a complete view of the image every 256 IF steps, resulting in a temporal resolution advantage for EL by a factor of n.

Despite these architectural and temporal differences, the performance gap between IF Coding and EL remains narrow, demonstrating the promising potential of IF Coding as an alternative, biologically inspired spike encoding method. As summ-arized in Table B, we not only visualized and analyzed spike sequences but also trained and evaluated four distinct SNN models using various spike-encoded inputs. The results confirm that IF Coding outperforms Rate and Temporal Coding in both accuracy and biological plausibility. Moreover, the proposed Artificial Visual Photoreceptor Layer, which increases input dimensionality by emulating both cone and rod cell functions, further enhances model accuracy and enables IF Coding to approach EL-level performance while preserving a biologically inspired structure. We thus conclude that leveraging the signal transformation capabilities of biological neuron models presents a promising research direction for future SNN development.

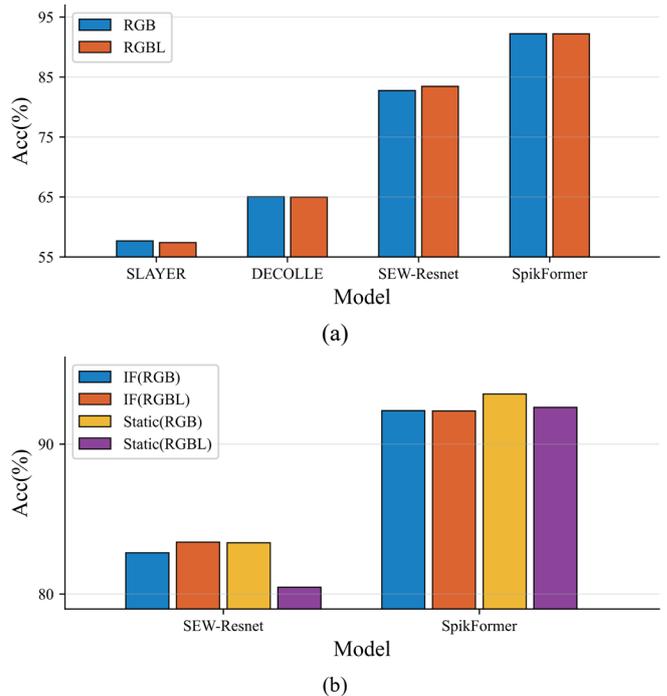

(a)

(b)

Figure 6. Accuracy of SNNs trained with color and color-luminance information. (a) Using IF coding. (b) Comparisons between IF coding and encoding layer.

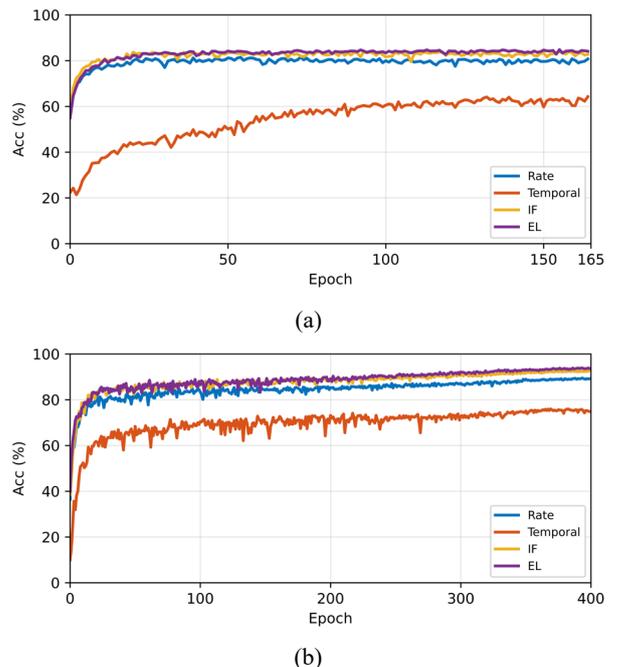

(a)

(b)

Figure 7. Validation curves on the CIFAR-10 dataset using RGB input: (a) SEW-ResNet, (b) SpikFormer.

## A. Color Space Model Substitution

To further investigate the characteristics of IF Coding, we conducted experiments by replacing the RGB color space of the image data with alternative color space models. Specifically, we evaluated LMS, LAB, and YUV color spaces, applying IF Coding to each and training SNN models accordingly to observe the resulting changes in classification accuracy.

Regarding the integration of these color spaces with IF Coding, the LMS color space is derived from the sensitivity of the three types of cone photoreceptors in the human retina, as discussed in Section II.C. These cones are sensitive to short (S), medium (M), and long (L) wavelengths of light, respectively. Thus, LMS color representation directly reflects the physiological response of the human visual system to incoming light, making it a biologically relevant choice for our experiments.

The LAB color space, also inspired by human visual perception, consists of three components: L (lightness), A (a chromatic component representing the green–red axis, with positive values indicating red and negative values indicating green), and B (representing the blue–yellow axis, with positive values indicating yellow and negative values indicating blue). This color model is widely used because it encompasses the full range of colors perceivable by the human eye.

The YUV color space is a commonly used model in image and video compression systems. It separates luminance (Y) from chrominance (U and V), where U and V represent blue and red color differences, respectively. This model is designed based on the observation that the human eye is more sensitive to brightness than color variations.

Notably, both LAB and YUV color spaces may contain negative values or values that fall outside the typical [0, 255] range used in standard image formats. To enable IF Coding, we applied value shifting and normalization to bring all components within a valid and consistent range.

The experimental results are summarized in Table C, and the accuracy performance across four SNN models is visualized in Fig. 8. With the exception of SLAYER's notably lower accuracy in the LAB and YUV color spaces, the models trained with RGB color space data consistently achieved slightly higher accuracy than those trained with data in the LMS, LAB, or YUV color spaces. Overall, differences in accuracy among the color spaces were relatively minor; however, the degraded performance of SLAYER when trained on LAB and YUV inputs suggests that these two color spaces may not be optimal for SNN training. In contrast, the commonly used RGB color space remains a robust and effective choice for SNN input representation.

## B. Comparison with Dynamic Event-Based Datasets

To further investigate whether the multidimensional spike data generated via neuron-based spike encoding and the artificial visual photoreceptor layer indeed carry richer information, we conducted a comparative study using the CIFAR10-DVS dataset, which consists of dynamic event data captured by recording static CIFAR-10 images through a motion-based differential platform. The comparison results are illustrated in Fig. 9, with detailed numerical values provided in Table D.

From Fig. 9, it is evident that although there is a significant difference in the number of time steps used, the event-based data, by design, only preserve temporal changes, and thus discard much of the static image features. This inherent sparsity in event representation leads to noticeably lower classification accuracy—approximately 10% lower on average—when compared to the performance achieved using IF Coding applied to static images.

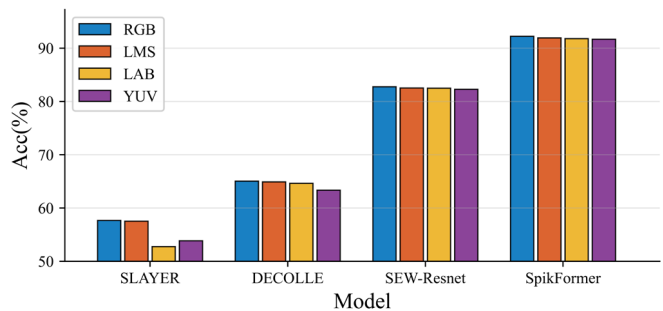

Figure 8. Accuracy comparison with different color spaces.

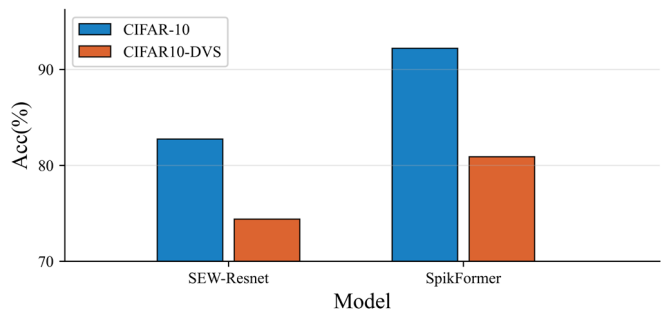

Figure 9. Accuracy between static image-based encodings and dynamic event-based dataset.

These results suggest that the multidimensional spike data generated via our biologically inspired encoding approach encapsulate more comprehensive visual information than that found in standard event-based data. Therefore, based on the observed accuracy differences, we conclude that neuron-modeled spike encoding, especially when combined with artificial photoreceptor structures, offers a richer and more informative representation than conventional event-driven datasets such as CIFAR10-DVS.

## C. Computational Resource Analysis

Given that SNNs, which are grounded in neuromorphic computing principles, are widely recognized for their low-power characteristics, it is essential to evaluate whether the proposed neuron-based spike encoding and artificial visual photoreceptor layer align with these energy-efficient objectives. To this end, we conducted a comprehensive analysis of FLOPs, parameter counts, and actual power consumption of the IF module, alongside two established models incorporating EL: SEW-ResNet and SpikFormer.

We begin by presenting the statistics for FLOPs and parameter counts. For consistency, we fixed the batch size, input data channel, and time step to 1, and set the input image size to 32×32 pixels, consistent with the CIFAR-10 dataset. The results are shown in Table I. In the case of the IF module, spike encoding involves only basic operations such as floating-point addition/subtraction and threshold comparison. It requires no parameter tuning or weight-based modulation, and no learning is involved. As a result, the FLOPs is merely 4096, and the parameter count is 0, indicating negligible computational overhead compared to SEW-ResNet and SpikFormer, which process information via convolutional layers. It is important to note that EL performs convolutional feature extraction before spike generation, leading to significantly higher FLOPs and nonzero parameters due to learnable convolution kernels.

Subsequently, we measured the actual power consumption during inference on the CIFAR-10 test set using the trained SEW-ResNet and SpikFormer models. In particular, we aimed to examine whether EL-based spike encoding consumes more energy compared to directly inputting spike-encoded data. To this end, we implemented a multi-threaded power monitoring utility. The experiments were conducted using batch size = 1 on a system equipped with an NVIDIA RTX 3060 GPU, the same environment used for training the DECOLLE model. The software stack consisted of NumPy v2.2.6 and PyTorch v2.7.

The power monitoring tool, based on the pynvml package, sampled GPU power every 1μs during model execution. Monitoring was activated immediately before model inference and deactivated immediately after, capturing fine-grained power usage and execution time. The experiment averaged power usage over 10,000 test images. First, we evaluated the power required solely for the encoding process, with results presented in Fig. 10. Next, we compared the power consumption of SEW-ResNet and SpikFormer when using different encoded input types, as shown in Fig. 11.

It is worth mentioning that all measurements were conducted without applying any software or hardware optimization, and thus some variance due to system overhead is expected. Nevertheless, based on computational principles, it is anticipated that EL, which includes convolution operations, will exhibit higher power demands.

A comparative analysis of computational resource consumption trends for different encoding schemes is illustrated in Fig. 10(a) and (b), representing processing time and power consumption, respectively. As shown, Rate Coding, Temporal Coding, and IF Coding exhibit generally comparable and relatively low requirements in both computation time and power usage, with IF Coding appearing slightly lower than the other two. In contrast, EL requires substantially higher processing time and power consumption. These trends are consistent with the conclusions derived from the detailed numerical analysis presented in Table E.

A comparative analysis of computational resource consumption for different encoding schemes applied to SEW-ResNet and SpikFormer is illustrated in Fig. 11(a) and (b), representing processing time and power consumption, respectively. As shown, the spike data obtained through Rate Coding, Temporal Coding, and the IF Layer lead to similar and relatively low requirements in both computation time and power usage. In contrast, the spike data generated via EL result in substantially higher processing time and power consumption compared to the aforementioned methods. These trends are consistent with the conclusions derived from the detailed numerical analysis presented in Table F.

As summarized in the detailed numerical results of Table F, the EL-based SEW-ResNet and SpikFormer models exhibit lower overall average computational resource usage compared to other encoding methods. This reduction, however, is primarily attributed to the small number of discrete time steps employed (6 for SEW-ResNet and 4 for SpikFormer). When normalized per time step, the resource usage is substantially higher, indicating that the convolutional operations required by EL are relatively power-intensive. Although SEW-ResNet and SpikFormer trained with static images using EL (as reported in Fig. 6(b)) achieve slightly higher accuracy, this improvement come-s at the cost of elevated per-step power consumption, thereby limiting their alignment with the low-power objectives fundamental to neuromorphic computing.

Table I
SUMMARY OF FLOPS AND PARAMETER COUNTS

| Module / Model | IF Module | SEW-Resnet | | Spikformer | |
|---|---|---|---|---|---|
| | | EL | Total | EL | Total |
| FLOPs | 4096 | 458.8 K | 383.6M | 1.08M | 1.87 G |
| Parameters | 0 | 193 | 768.9 K | 528 | 9.33 M |

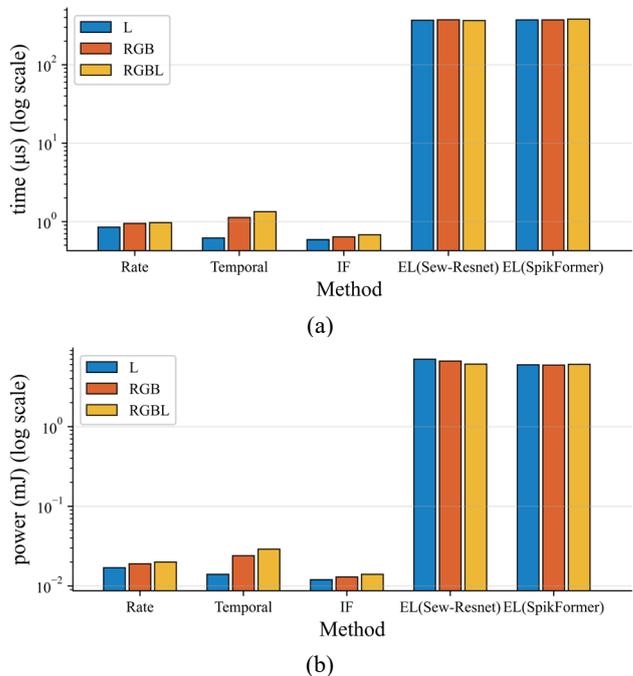

Figure 10. Comparative analysis of resources required for different encoding methods: (a) processing time and (b) power consumption.

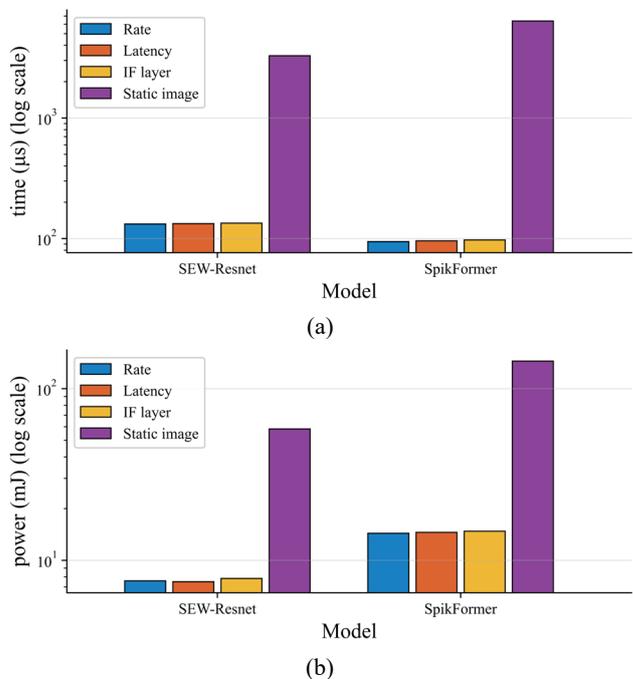

Figure 11. Comparative analysis of resources required for SNNs when tested with different data types: (a) processing time and (b) power consumption.

## V. CONCLUSION AND FUTURE WORK

This study proposes the concept of neuron-inspired spike encoding, derived from the functional characteristics of the human visual system. Instead of traditional Rate Coding and Temporal Coding, which are primarily used to analyze spike regularity, we adopt a biologically plausible spike encoding approach based on the intrinsic firing behavior of neurons. Additionally, we design an artificial visual photoreceptor layer inspired by retinal cone and rod cells, enabling artificial neurons to generate corresponding color and luminance spike signals. This design aims to maximize the informational content carried by spike trains.

In our experiments, we employed the IF neuron model to perform IF Coding on static images. By incorporating the artificial visual photoreceptor layer to generate color and luminance spike streams, we trained multiple SNNs and demonstrated that the spike data produced through IF Coding effectively convey richer information. The superior performance in testing accuracy confirms the feasibility and effectiveness of both the neuron-based spike encoding and the artificial photoreceptor layer. Furthermore, power consumption analysis shows that this encoding approach aligns with the low-power principles of neuromorphic computing, reaffirming its advantages in energy-efficient design.

It is important to note that, in our implementation, the firing threshold of the IF neurons was uniformly set to 1. We did not perform threshold tuning or adjust image resolution to match the density and distribution of biological photoreceptors. This decision was primarily constrained by current computational methods and hardware limitations, which led us to conclude that such adjustments might not yield significant benefits at this stage. However, we hypothesize that carefully adjusting the firing threshold and input resolution to better mimic biological photoreceptor characteristics could potentially enhance the SNN's learning capability and overall performance. Additionally, since our experiments were based on the fixed assumptions outlined in Section III.A, we did not vary the time step parameter, and thus did not explore how the temporal length of IF Coding might affect accuracy or learning dynamics—this remains an open question for future investigation.

Moreover, to isolate the impact of different spike encoding methods, we designed the experiments to minimize changes to the original SNN architectures. This was to avoid potential confounding effects such as impaired convergence or degraded learning performance. Ideally, the Neuron-like Encoding Module should be implemented as a dedicated Neuron-like Encoding Layer, directly serving as the input layer of the SNN. Under the assumption of stable illumination during an extremely short exposure time, static images could be continuously and directly streamed into the SNN, allowing the encoding layer to convert optical energy into spike signals in real time—as demonstrated in Table II under the power consumption analysis in Section IV.B.

In conclusion, our work successfully enhances the informational richness of spike-encoded data within the framework of neuromorphic efficiency, enabling SNNs to learn more effectively while revealing the importance of temporal resolution in neural encoding. We hope that the proposed concept of neuron-inspired spike encoding and the design of an artificial visual photoreceptor layer will stimulate further research, helping SNNs overcome current limitations and expand their applicability across broader domains.